\RequirePackage[T1]{fontenc}
\documentclass[conference]{IEEEtran}
\IEEEoverridecommandlockouts
\usepackage{cite}
\usepackage{amsmath,amssymb,amsfonts}
\usepackage{algorithmic}
\usepackage{graphicx}
\usepackage{multirow}
\usepackage{booktabs}
\usepackage{placeins}
\usepackage{stfloats}
\usepackage{url}
\usepackage{textcomp}
\usepackage[table]{xcolor}
\usepackage{pifont}
\usepackage[colorlinks=true, urlcolor=blue, linkcolor=blue, citecolor=blue]{hyperref}
\def\BibTeX{{\rm B\kern-.05em{\sc i\kern-.025em b}\kern-.08em
    T\kern-.1667em\lower.7ex\hbox{E}\kern-.125emX}}
\begin{document}

\title{DobotTrack: Target-Grounded Vision-Language-Action Learning for Embodied Person Tracking}

\author{
\IEEEauthorblockN{
\shortstack[c]{
Shengjun Deng\textsuperscript{1},
Yinghan Zhang\textsuperscript{2,*},
Chengrui Wei\textsuperscript{3},
Liu Liang\textsuperscript{3},
Songyuan Liu\textsuperscript{3}
\\[3pt]
Pengkun Zheng\textsuperscript{3,\dag},
Yajun Zheng\textsuperscript{3},
Ying Zhang\textsuperscript{3},
Yuan Tian\textsuperscript{3},
Xi Li\textsuperscript{3},
Teng Wang\textsuperscript{4*}
}
}
\IEEEauthorblockA{
\textsuperscript{1}Foshan University \qquad
\textsuperscript{2}Tsinghua University\\
\textsuperscript{3}Dobot Robotics \qquad
\textsuperscript{4}The University of Hong Kong\\[2pt]
\textsuperscript{*}Corresponding authors:
tengwang@connect.hku.hk,
yinghan-24@mails.tsinghua.edu.cn\\
\textsuperscript{\dag}Project Lead
}
}

\maketitle

\begin{figure*}[t]
    \centering
    \includegraphics[width=\textwidth]{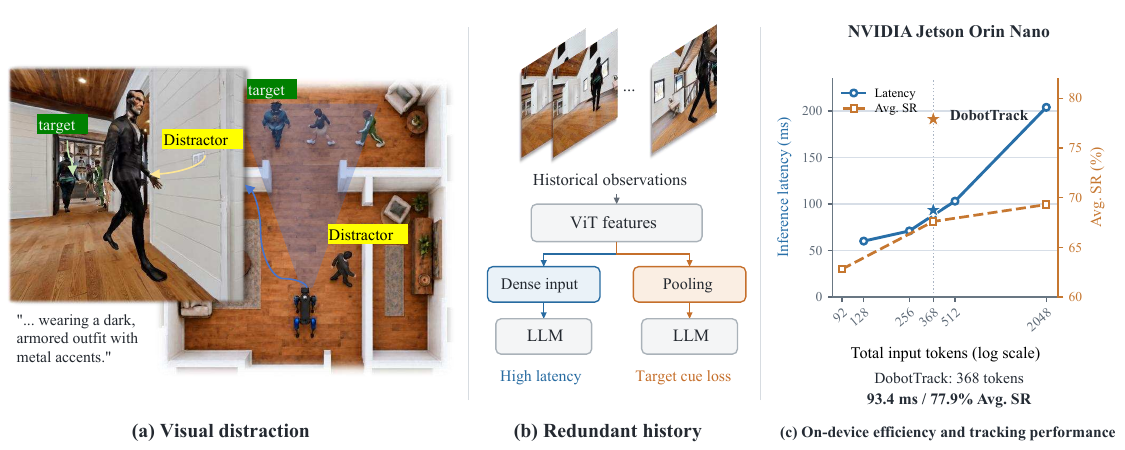}
    \caption{\textbf{Challenges in language-guided embodied person tracking.}
    \textbf{(a)} Distractors and occlusion complicate target grounding in the robot's limited field of view.
    \textbf{(b)} Dense visual histories increase inference cost, while indiscriminate pooling can discard target-specific cues.
    \textbf{(c)} Median inference latency on an NVIDIA Jetson Orin Nano (FP16, batch size 1, including the visual encoder; 90 measurements per point) increases with the total input token count.
    Reducing the token count from 2048 to 256 lowers latency by approximately 52\%, motivating target-conditioned history compression for on-device deployment.}
    \label{fig:tracking_overview}
\end{figure*}


\begin{abstract}
Language-guided embodied person tracking requires a mobile agent to identify a target from a referring description and consistently follow the target under egocentric observations. We present \textbf{DobotTrack}, a target-grounded vision-language-action (VLA) framework that integrates auxiliary grounding supervision with efficient temporal modeling. We systematically investigate five auxiliary tasks---person segmentation, bounding-box grounding, visual-spatial classification (VSC), future DINO feature prediction, and future-frame prediction---and adopt complementary mask and bounding-box supervision to enhance target delineation and localization. We introduce Q-identity, a target-conditioned historical visual compression module that leverages target-grounding predictions to selectively preserve identity-discriminative appearance features and spatial cues in compact historical tokens. On EVT-Bench, our 0.8B-parameter model achieves an average success rate of 77.9\%, exceeding OmTrackVLA by 16.9 percentage points across the three tasks. We further construct a real-world person-tracking dataset collected using a quadruped robot, comprising 126 episodes and 596,323 mask-annotated frames. The target masks enable automatic derivation of bounding boxes and visual-spatial labels encoding horizontal position and apparent image scale, providing additional grounding supervision without separate manual annotation of these labels. Together, our framework and annotation pipeline support learning target-grounded tracking policies from simulated and real-world data. Code is available at \url{https://github.com/Dobot-Team/dobot_track}.
\end{abstract}

\section{Introduction}
\label{sec:introduction}

Tracking a person specified in natural language requires an embodied agent to determine whom to track while continually deciding how to move. Unlike tracking in a recorded video, the agent's actions change its viewpoint and the target's visibility, coupling visual recognition with motion generation~\cite{maalouf2023followanything,rspt}. Appearance variation, occlusion, and nearby distractors make this coupling particularly important: a plausible trajectory is insufficient if the agent tracks the wrong person. Language-guided embodied tracking therefore calls for representations that connect the referring description, the intended target's image location and the agent's future motion.

Vision--language--action models provide a way to learn these capabilities within a shared model. Uni-NaVid unifies several embodied navigation tasks, including person tracking~\cite{zhang2024uninavid}, while TrackVLA shares a language-model backbone between target recognition and trajectory planning~\cite{wang2025trackvla}. Subsequent work introduces explicit spatial reasoning, target-identification memory, and image-grounded referring decisions~\cite{liu2025trackvlaplusplus,ye2026refertrack}. During sustained tracking, changes in illumination and viewpoint can substantially alter the same person's visual appearance, making target association and re-identification after occlusion difficult~\cite{zeng2020illumination,personreid}. These variations motivate retaining target-discriminative information across observations. These challenges, illustrated in Fig.~\ref{fig:tracking_overview}, raise two complementary questions. First, \emph{how can a tracking policy be encouraged to represent the specified target explicitly, rather than relying on trajectory imitation alone?} Second, \emph{how can historical visual observations be efficiently compressed while preserving target-discriminative appearance cues and scene context essential for reliable person tracking?}

We address these questions through target-centric perceptual supervision and identity-aware history compression. We present \textbf{DobotTrack}, a target-grounded VLA framework built on InternVL3~\cite{internvl3}. We systematically investigate person segmentation, bounding-box grounding, visual-spatial classification, future DINO feature prediction, and future-frame prediction as auxiliary tasks to improve tracking policy learning. The final model combines mask and box supervision to couple fine-grained foreground delineation with explicit target localization. Building on this grounding, Q-identity compresses historical observations into compact tokens that retain identity-discriminative appearance and spatial context. The language model captures temporal dependencies for trajectory prediction from the chronological sequence of compressed historical and current-frame tokens.

The model processes an appearance-based instruction together with a temporal window of RGB observations. The learned mask token and box token precede the waypoint tokens in the causal sequence, allowing the waypoint states to attend to their target-related representations. Task-specific heads decode the resulting hidden states into the target mask, bounding box, and future trajectory in one forward pass.

A second consideration is the alignment between policy-learning objectives and real-world annotations. EVT-Bench supports simulated embodied-tracking research~\cite{wang2025trackvla}, and TPT-Bench provides robot-egocentric tracking sequences with target boxes and synchronized sensor streams, including odometry~\cite{ye2025tptbench}. Complementing these resources, we construct a quadruped-collected dataset with 126 tracking episodes and 596,323 mask-annotated RGB frames. The recordings combine target descriptions, visual observations, and robot motion information. Target masks yield boxes and structured VSC labels, while temporally continuous recordings provide observation histories and future waypoint supervision. Episode-level partitioning keeps samples from the same recorded episode within a single data subset.

This report makes three contributions:
\begin{itemize}
    \item \textbf{A target-grounded tracking framework with systematic auxiliary supervision analysis.}
    We develop DobotTrack, a vision-language-action framework for embodied person tracking, and systematically investigate five auxiliary tasks: person segmentation, bounding-box grounding, visual-spatial classification, future DINO feature prediction, and future-frame prediction. Based on our empirical findings, we adopt complementary mask and box supervision to improve target delineation and spatial localization.

    \item \textbf{Identity-aware historical visual compression.}
    We introduce Q-identity, a target-conditioned historical visual compression module that leverages target-grounding predictions to preserve identity-discriminative appearance information and essential spatial cues in compact visual tokens, providing target-aware temporal context for subsequent trajectory prediction.

    \item \textbf{A real-world tracking dataset and annotation pipeline.}
    We construct a quadruped-collected person-tracking dataset comprising 126 episodes and 596,323 mask-annotated frames. We develop an annotation pipeline that derives bounding boxes and structured VSC labels for horizontal target position and apparent image scale from target masks, and organize the data into temporally structured learning samples with episode-level partitioning.
\end{itemize}

\section{Related Work}
\label{sec:related_work}

\subsection{Language-Guided Embodied Visual Tracking}

Embodied visual tracking couples target perception with active motion control. Follow Anything integrates open-set detection and tracking into a robotic following system~\cite{maalouf2023followanything}. RSPT learns a structure-aware representation by reconstructing the surroundings and predicting target motion~\cite{rspt}, while Zhong et al. combine visual foundation models for target segmentation with offline reinforcement learning for control~\cite{evt}. These methods establish the importance of target-aware representations for active tracking.

Recent VLA approaches learn language-conditioned perception and action within shared models. Uni-NaVid supports multiple navigation tasks, including person tracking~\cite{zhang2024uninavid}. TrackVLA couples recognition-oriented language modeling with an anchor-based diffusion trajectory decoder~\cite{wang2025trackvla}. TrackVLA++ introduces Polar-CoT spatial reasoning and a gated Target Identification Memory~\cite{liu2025trackvlaplusplus}. ReferTrack selects the referred target from indexed candidate boxes, predicts waypoints conditioned on this decision, and injects historical box information into visual context~\cite{ye2026refertrack}. Our work systematically examines target-centric auxiliary tasks for tracking policy learning and combines mask supervision for learning target representations with box supervision for spatial localization. We further investigate how target-grounding predictions can guide historical visual compression for subsequent action prediction.

\subsection{Visual Grounding and Target-Centric Supervision}

Visual grounding associates language with image regions. Grounding DINO uses language-conditioned visual representations for open-set detection~\cite{groundingdino}, while Kosmos-2 represents grounded text with location tokens~\cite{kosmos2}. At pixel resolution, SAM 2 supports promptable segmentation in images and videos~\cite{sam2}. LISA connects a segmentation token in a multimodal language model to mask prediction through an embedding-as-mask formulation~\cite{lisa}. These works provide relevant foundations for language-conditioned localization and segmentation; token-conditioned mask decoding is not specific to embodied tracking.

DobotTrack uses grounding as supervision for a tracking policy, rather than treating localization as the final task. Mask prediction and box regression supervise the spatial extent of the referred person, while waypoint regression supervises the agent's motion. The real-world structured VSC labels further express box-derived position and scale as categorical labels. Because these annotations originate from the same masks, their role is to provide different learning objectives over shared target information, not to introduce independent geometric measurements. We compare these auxiliary objectives under a shared tracking framework and combine mask and box supervision in the final model.




\subsection{Identity-Preserving Temporal Memory}

Maintaining target identity over time is essential for person tracking, particularly when pedestrians cross paths, undergo appearance changes, or temporarily disappear. Visual tracking and video segmentation provide relevant mechanisms for temporal association. Online person re-identification~\cite{personreid} addresses appearance changes and distractors during robot following. MOTR~\cite{motr} propagates and updates track queries across frames to represent individual objects over time. XMem~\cite{xmem} combines memory stores operating at different temporal scales and consolidates working-memory features into compact long-term representations. SAM 2~\cite{sam2} uses streaming memory to support object segmentation in videos. While these methods primarily address temporal consistency at the perception level, embodied tracking additionally requires historical target information to inform the agent's movement.

Recent embodied models incorporate several forms of temporal memory. TrackVLA++~\cite{liu2025trackvlaplusplus} maintains target identity through gated memory updates, ReferTrack~\cite{ye2026refertrack} embeds historical target boxes into visual context, and VLingNav~\cite{vlingnav} combines linguistic summaries with visual features in persistent memory. These developments motivate further investigation into which information should be retained under a limited memory budget. For long-video understanding, LLaMA-VID~\cite{llamavid} represents each frame with compact context and content tokens, while LongVU~\cite{longvu} adaptively reduces spatial and temporal visual redundancy. BLIP-2~\cite{blip2} provides a query-based mechanism, Q-Former, for extracting compact visual representations. Building on query-based compression, Q-identity places target conditioning within the formation of compact historical visual tokens. Target-region features condition learned compression queries, which also aggregate full-image context under a fixed token budget. The module is designed to retain target appearance and spatial cues together with scene context, while the language model performs temporal reasoning over the chronological sequence of compressed frames.

\subsection{Real-World Data for Person Tracking}

The data associated with TrackVLA and EVT-Bench support large-scale simulated tracking and recognition learning~\cite{wang2025trackvla}. TPT-Bench complements simulation with long-term, robot-egocentric target-person tracking sequences collected using a sensor-equipped cart~\cite{ye2025tptbench}. In addition to target boxes, it includes odometry and other synchronized modalities, making it relevant to both temporal perception and robot-following research.

Our dataset is collected with a quadruped and is organized around sequence-level VLA supervision: observation histories and target descriptions are paired with future robot waypoints and current-frame grounding annotations.

\section{Method}
\subsection{Problem Formulation and Overview}
\label{sec:problem_overview}

\noindent{\fontfamily{lmr}\bfseries\selectfont Problem formulation.}
We consider language-guided embodied person tracking~\cite{wang2025trackvla,ye2026refertrack}, where a mobile agent follows a person identified by an appearance-based language instruction $\ell$ using a forward-facing RGB camera. At each time step $t$, the agent receives a temporal window of $H=8$ RGB observations sampled at $5$Hz, denoted by $\mathcal{O}_t=(o_{t-7\Delta},\ldots,o_{t-\Delta},o_t)$, where $\Delta=0.2$\,s is the observation sampling interval. Given $\ell$ and $\mathcal{O}_t$, the agent predicts a sequence of $P=8$ future waypoints, $\hat{\mathbf{W}}=\{\hat{\mathbf{w}}_p\}_{p=1}^{P}$, at a frequency of $10$Hz, yielding a prediction horizon of $0.8$s. Each waypoint $\hat{\mathbf{w}}_p=(\hat{x}_p,\hat{y}_p,\hat{\theta}_p)$ specifies the agent's future planar pose in its current egocentric coordinate frame.

\begin{figure*}[t]
    \centering
    \includegraphics[width=\textwidth]{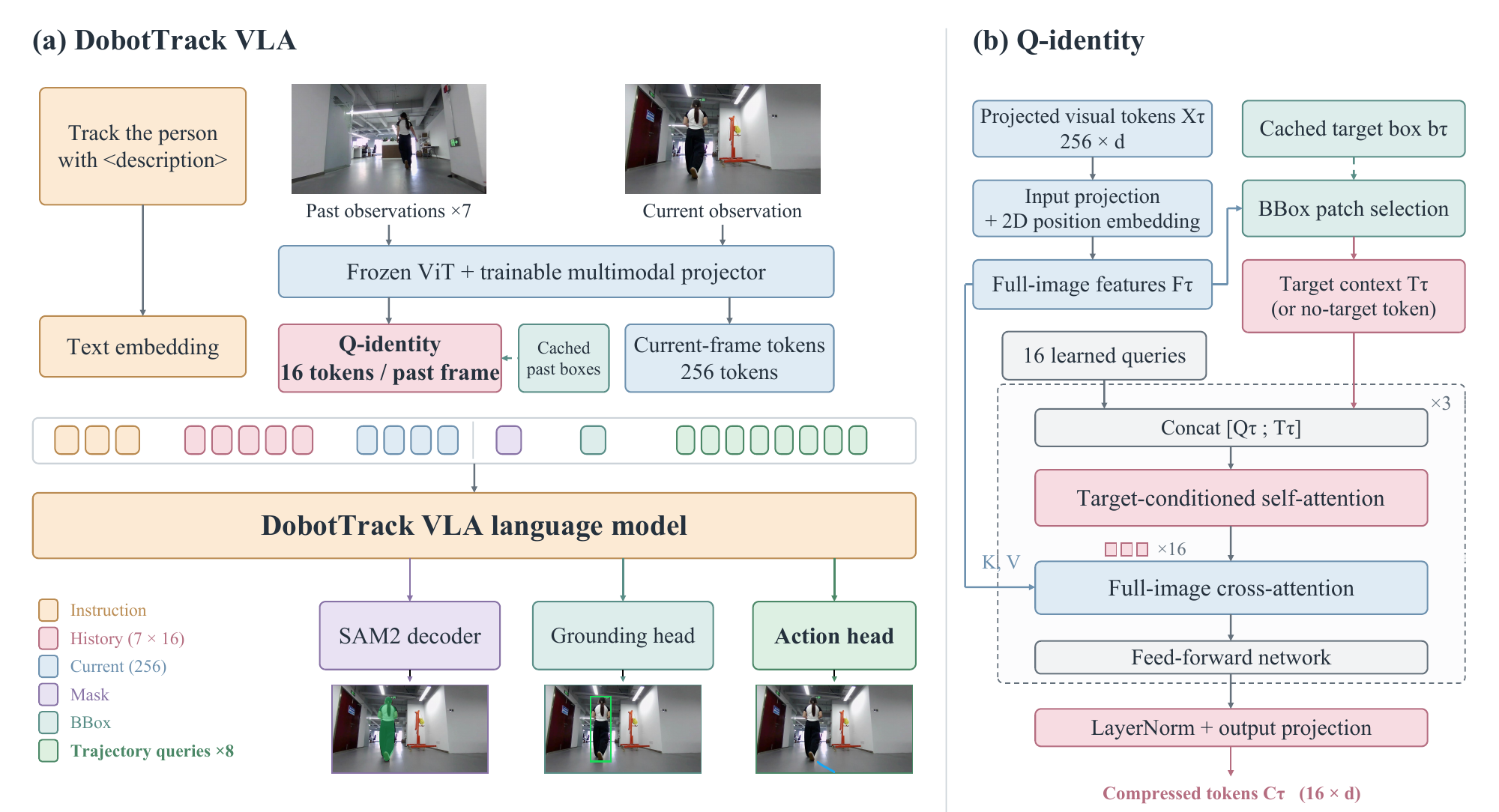}
    \caption{Overview of DobotTrack and Q-identity. Left: the VLA processes a target description, current-frame visual tokens, and compressed historical tokens to predict target grounding and future waypoints. Target-grounding predictions guide historical visual compression for subsequent decisions. Right: Q-identity aggregates target-region features and full-image context through joint query--target self-attention, full-image cross-attention, and feed-forward updates. The resulting compact historical tokens are fed back into the VLA's visual input.}
    \label{fig:architecture}
\end{figure*}

\noindent{\fontfamily{lmr}\bfseries\selectfont Model overview.}
As shown in Fig.~\ref{fig:architecture}, our model builds on the InternVL3~\cite{internvl3} vision-language backbone. The visual encoder and multimodal projector map each RGB observation $o_t$ to visual tokens $\mathbf{X}_t\in\mathbb{R}^{N\times d}$, where $N$ is the number of visual tokens per frame and $d$ is the language-model embedding dimension. Q-identity compresses historical tokens by jointly exploiting target-region cues and full-image context. The compressed representations are concatenated chronologically with current-frame tokens and processed with the instruction $\ell$ by the language model to integrate temporal and linguistic information. A learned mask token and a bounding-box grounding token precede $P$ waypoint tokens, in that order, allowing trajectory prediction to attend to both target-supervised representations through causal attention. Task-specific heads jointly predict the current target mask, bounding box, visibility, and future egocentric waypoints in a single forward pass. During online inference, cached historical box and visibility predictions guide target-region selection for subsequent visual history compression.

\subsection{Target-Centric Supervision and Trajectory Prediction}
\label{sec:joint_decoding}

We systematically investigate five auxiliary tasks: person segmentation, bounding-box grounding, visual-spatial classification, future DINO feature prediction, and future-frame prediction. The final model jointly learns person segmentation, bounding-box grounding, and trajectory prediction. Segmentation supervision delineates the target representation, while box supervision captures its overall location and extent. These complementary objectives encourage the shared backbone to capture the target's appearance and spatial extent for trajectory prediction. In this section, we focus on two auxiliary tasks: person segmentation and bounding-box grounding. The remaining auxiliary tasks are introduced and comparatively evaluated in the ablation study section.


\noindent{\fontfamily{lmr}\bfseries\selectfont Person segmentation.}
Let $\mathbf{h}_{\mathrm{mask}}$ denote the contextualized
hidden state of the mask token. The segmentation head
combines this state with the dense current-frame features
to predict the referred person's foreground probability map:
\begin{equation}
    \hat{m}_t
    =
    \sigma\!\left(
        f_{\mathrm{mask}}
        (\mathbf{X}_t,\mathbf{h}_{\mathrm{mask}})
    \right),
    \label{eq:mask_prediction}
\end{equation}
where $f_{\mathrm{mask}}$ consists of trainable visual
and prompt projections followed by a pretrained SAM2
mask decoder~\cite{sam2}, and $\sigma$ denotes the
sigmoid function. The projections map the dense
current-frame features and the mask-token hidden state
into image embeddings and a sparse prompt, respectively. This branch provides
explicit pixel-level supervision for the target
representation used by subsequent waypoint tokens.

\noindent{\fontfamily{lmr}\bfseries\selectfont Bounding-box grounding.}
For the bounding-box grounding task, a box head maps the grounding token's hidden state $\mathbf h_{\mathrm{box}}$ to normalized image-space coordinates:
\begin{equation}
\hat{\mathbf b}_t=f_{\mathrm{box}}(\mathbf h_{\mathrm{box}}),
\label{eq:box_prediction}
\end{equation}
where $\hat{\mathbf b}_t=(\hat{c}_x,\hat{c}_y,\hat{w},\hat{h})$ represents the normalized box center, width, and height. The head also predicts target visibility. This task supervises the referred person's location and spatial extent.


\noindent{\fontfamily{lmr}\bfseries\selectfont Trajectory prediction.}
The action head maps the hidden state $\mathbf h_p$ of the $p$-th waypoint token to its corresponding egocentric waypoint:
\begin{equation}
\hat{\mathbf w}_p=f_{\mathrm{act}}(\mathbf h_p),
\qquad p=1,\ldots,P.
\label{eq:action_head}
\end{equation}
The current target mask, bounding box, and future waypoints are obtained in the same forward pass.

\subsection{Q-identity: Target-Conditioned Visual History Compression}
\label{sec:history_compression}

Visual history supports target association and motion reasoning, but
retaining dense tokens from every frame increases the language model's
input length. Motivated by compact video representations
\cite{llamavid,longvu} and query-based compression~\cite{blip2},
we introduce Q-identity, which compresses each historical frame into
$N_q$ target-conditioned tokens. The compressor first incorporates
localized target features through target--conditioned self-attention,
then aggregates full-image features through cross-attention. This
design gives the compressed representation access to both target
appearance and location cues and the scene context needed for planning.

\noindent\textbf{Target context construction.}
Let $\tau$ index a historical frame, and let
$\mathbf X_\tau\in\mathbb R^{N\times d}$ denote its
visual features after the visual encoder and multimodal
projector. We linearly project these features into
$d_c$ dimensions and add spatial embeddings derived
from normalized 2D grid coordinates, obtaining
full-image features
$\mathbf F_\tau\in\mathbb R^{N\times d_c}$.

During training, target regions are determined from
perturbed ground-truth boxes; at inference, we use
cached box and visibility predictions from earlier
decisions. The visual tokens corresponding to each
selected region form the target context $\mathbf T_\tau$,
providing target-specific appearance and spatial cues,
while $\mathbf F_\tau$ retains the surrounding scene
information. When no valid target region is available,
we use a learned no-target token.

\noindent\textbf{Target-conditioned aggregation.}
Each historical frame initializes $N_q$ query tokens from the same learned embeddings, which are updated independently through successive attention blocks. Within each block, the
query states $\mathbf Z_\tau\in\mathbb R^{N_q\times d_c}$ are concatenated with
target-region features to form
$\mathbf U_\tau=[\mathbf Z_\tau;\mathbf T_\tau]$.
Linear projections of the normalized $\mathbf U_\tau$
produce queries $\mathbf Q_\tau^{\mathrm{tgt}}$, keys
$\mathbf K_\tau^{\mathrm{tgt}}$, and values
$\mathbf V_\tau^{\mathrm{tgt}}$ for target-conditioned
self-attention:
\begin{equation}
\mathbf A_\tau^{\mathrm{tgt}}=
\operatorname{softmax}\!\left(
\frac{\mathbf Q_\tau^{\mathrm{tgt}}(\mathbf K_\tau^{\mathrm{tgt}})^\top}
{\sqrt{d_h}}
\right)\mathbf V_\tau^{\mathrm{tgt}}.
\label{eq:history_joint_attention}
\end{equation}
Only the first $N_q$ positions of
$\mathbf A_\tau^{\mathrm{tgt}}$ are retained to update
the query states through a residual connection;
target-region features remain fixed across blocks.
The updated query states $\mathbf Z_\tau$ are normalized and linearly projected into queries \(\mathbf Q_\tau^{\mathrm{ctx}}\), while the full-image features \(\mathbf F_\tau\) are projected into keys \(\mathbf K_\tau^{\mathrm{ctx}}\) and values \(\mathbf V_\tau^{\mathrm{ctx}}\) for cross-attention:
\begin{equation}
\mathbf A_\tau^{\mathrm{ctx}}=
\operatorname{softmax}\left(
\frac{\mathbf Q_\tau^{\mathrm{ctx}}(\mathbf K_\tau^{\mathrm{ctx}})^\top}{\sqrt{d_h}}
\right)\mathbf V_\tau^{\mathrm{ctx}}.
\label{eq:history_cross_attention}
\end{equation}
Here, $d_h$ denotes the per-head dimension, and the equations omit the multi-head formulation for clarity. 
These two attention stages integrate target-specific cues and global scene context into the learned queries. Residual
connections and a feed-forward network complete each
block. After three blocks, normalization and projection
yield compressed tokens
$\mathbf C_\tau\in\mathbb R^{N_q\times d}$.

We use $N_q=16$, $d_c=512$, and eight attention heads.
Frames are compressed independently with shared
parameters, and their outputs are concatenated
chronologically with the current-frame tokens for
temporal reasoning by the language model. For a window
of eight frames, including the current frame, the
visual-token count decreases from
$8\times256=2048$ to $7\times16+256=368$.

\subsection{Training Objective and Online Inference}
\label{sec:training_inference}

The model jointly learns trajectory prediction, current-frame person segmentation, and bounding-box grounding:
\begin{equation}
\mathcal L =
\mathcal L_{\mathrm{traj}}
+ \lambda_{\mathrm{seg}}\mathcal L_{\mathrm{seg}}
+ \lambda_{\mathrm{box}}\mathcal L_{\mathrm{box}},
\label{eq:joint_loss}
\end{equation}
where $\lambda_{\mathrm{seg}}=0.001$ and
$\lambda_{\mathrm{box}}=0.01$.

\noindent\textbf{Trajectory prediction.}
For ground-truth waypoints
$\mathbf w_p=(x_p,y_p,\theta_p)$, the trajectory objective is
\begin{equation}
\mathcal L_{\mathrm{traj}}
=
\frac{1}{P}\sum_{p=1}^{P}
\sum_{k\in\{x,y,\theta\}}
\alpha_k
\left(\hat w_{p,k}-w_{p,k}\right)^2,
\label{eq:trajectory_loss}
\end{equation}
with $(\alpha_x,\alpha_y,\alpha_\theta)=(1,3,2)$.
Here, $\theta$ denotes yaw.

\noindent\textbf{Person segmentation.}
The mask objective combines segmentation supervision with
intersection over union (IoU) score regression and object-presence classification:
\begin{equation}
\mathcal L_{\mathrm{seg}}
=
\mathcal L_{\mathrm{BCE}}
+\mathcal L_{\mathrm{Dice}}
+\mathcal L_{\mathrm{IoU\text{-}score}}
+\mathcal L_{\mathrm{obj}}.
\label{eq:mask_loss}
\end{equation}
Predicted mask logits are bilinearly resized to the ground-truth
resolution before computing pixelwise binary cross-entropy and
Dice loss~\cite{vnet}. Dice loss uses a smoothing constant of $1$.
The IoU-score term is a mean squared error against the detached
IoU of the thresholded predicted mask and ground truth.
The object-presence term is binary cross-entropy, with a positive
label when the ground-truth mask is nonempty.

\noindent\textbf{Bounding-box grounding.}
The grounding head predicts a normalized bounding box
$(c_x,c_y,w,h)$ and target visibility. Its objective is
\begin{equation}
\mathcal L_{\mathrm{box}}
=
\beta_{\mathrm{reg}}\mathcal L_{\mathrm{SmoothL1}}
+\beta_{\mathrm{iou}}\mathcal L_{\mathrm{GIoU}}
+\beta_{\mathrm{vis}}\mathcal L_{\mathrm{vis}}.
\label{eq:box_loss}
\end{equation}
The loss weights are set to
$(\beta_{\mathrm{reg}},\beta_{\mathrm{iou}},
\beta_{\mathrm{vis}})=(5,2,1)$.
Coordinate regression and generalized IoU loss are
averaged over samples with visible targets. Visibility
is supervised using binary cross-entropy across the
batch, with labels determined by whether the target
mask is nonempty. Trajectory and mask objectives are
averaged over the batch.

We freeze the visual encoder and jointly optimize the multimodal
projector, language model, Q-identity, mask head, grounding head,
and action head. During training, input boxes are perturbed with
probability $0.8$, using center offsets proportional to box size
and multiplicative size perturbations with scale $0.1$.
We also compress the current frame with probability $0.3$, using
a box derived from its ground-truth mask. The mask decoder always
receives the uncompressed current-frame features.


\noindent\textbf{Online inference.}
At each decision step, the model receives the instruction and an
updated RGB window. Past frames are compressed using their cached
bounding-box predictions, while the current frame retains its
dense visual tokens. One forward pass predicts future waypoints,
the current target mask, bounding box, and visibility.
The current box and visibility prediction are then cached for use
when that observation becomes a historical frame in a later decision.
Predictions with visibility below $0.5$, and frames without a cached
prediction, use the no-target fallback. This ordering ensures that
compression uses only predictions available before the current
decision.

\section{Dataset}
  \label{sec:dataset}

\begin{figure*}[!t]
      \centering
      \includegraphics[width=\textwidth]{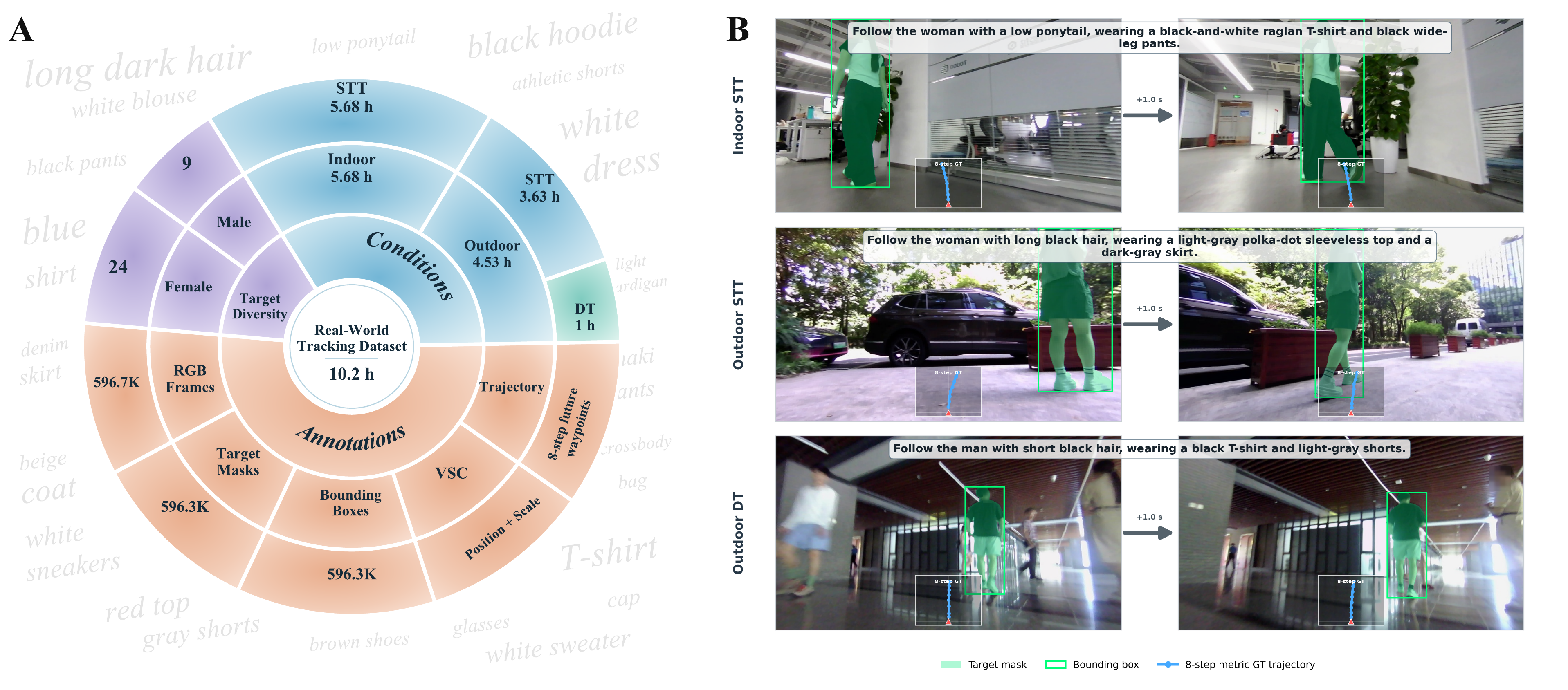}
      \caption{
      \textbf{Overview of the real-world tracking dataset.}
      \textbf{(A)} Dataset composition.
      The dataset contains approximately 10.2 hours of real-world human-tracking data collected in indoor and outdoor
      environments under both STT and DT conditions.
      The annotations include RGB observations, target masks, bounding boxes, structured VSC labels, and future robot
      trajectories.
      The dataset contains 33 target appearance profiles, including 24 female and 9 male profiles.
      \textbf{(B)} Representative qualitative examples from indoor STT, outdoor STT, and outdoor DT sequences.
      Each example shows the natural-language instruction describing the target and two frames from the same tracking sequence.
      Green overlays denote the target segmentation masks, green rectangles indicate the corresponding bounding boxes,
      and the blue curve visualizes the 8-step ground-truth future robot trajectory.
      }
      \label{fig:real_world_dataset_overview}
  \end{figure*}

  Our training data combine simulation data with a real-world robot-tracking dataset.
  The two sources are complementary: simulation provides large-scale and controllable supervision over diverse tracking
  conditions, while the real-world data expose the model to realistic human appearances, illumination changes, occlusions,
  distractors and camera motion.
  Both sources provide trajectory, segmentation-mask, and bounding-box annotations, along with structured visual-spatial classification labels. These annotations and labels support multi-task VLA training.

\subsection{Simulation Dataset}
\label{sec:simulation_dataset}

\noindent{\fontfamily{lmr}\bfseries\selectfont Data source.}
Our simulation data are derived from EVT-Bench~\cite{wang2025trackvla}, which is built on Habitat 3.0 using indoor
scenes from HM3D and Matterport3D.
Animated human avatars serve as tracking targets and distractors.
We use the three tracking settings defined by EVT-Bench: Single-Target Tracking (STT), Distracted Tracking (DT),
and Ambiguous Tracking (AT).
The resulting simulation training set contains approximately 1.8 million sequence samples with an STT:DT:AT ratio
of $1:2:1$.
Each sample provides forward-facing RGB observations, an appearance-based target instruction, an egocentric expert
trajectory, and target-centric auxiliary annotations.

\noindent{\fontfamily{lmr}\bfseries\selectfont Target-centric annotation.}
Target masks are extracted from simulator panoptic observations using the semantic identifier of the tracked person.
Bounding boxes are the tightest axis-aligned rectangles enclosing visible foreground masks, normalized as $(c_x,c_y,w,h)$.
Bounding boxes are used to derive left/center/right and small/medium/large VSC labels, using horizontal center thresholds of 1/3 and 2/3 and normalized area thresholds of 0.05 and 0.20, respectively.
Valid metadata yield near/medium/far distance labels at $1.5$ and $3.0$\,m. Binary visibility labels indicate whether the target is visible in the current camera view.

\begin{table*}[!t]
\centering
\caption{EVT-Bench results (\%). Multi-camera methods are external references and excluded from single-view ranking. Results and protocol labels follow the cited papers and the comparisons in ReferTrack and CoMaTrack. RL: \ding{51} indicates reinforcement learning, \ding{55} indicates no reinforcement learning; dashes denote unavailable entries. $^{*}$: co-trained with general navigation data; $^{\dagger}$: GroundingDINO detector; $^{\ddagger}$: SoM + GPT-4o perception. DobotTrack uses mask and box supervision, simulation and real-world training data, and 16 historical tokens per frame.}
\label{tab:main_results}
\footnotesize
\setlength{\tabcolsep}{2.7pt}
\renewcommand{\arraystretch}{1.03}
\begin{tabular}{lccrrr@{\hspace{10pt}}rrr@{\hspace{10pt}}rrr}
\hline
\multirow{2}{*}{Method} & \multirow{2}{*}{Size} & \multirow{2}{*}{RL} & \multicolumn{3}{c}{STT} & \multicolumn{3}{c}{DT} & \multicolumn{3}{c}{AT} \\
\cmidrule(lr){4-6}\cmidrule(lr){7-9}\cmidrule(lr){10-12}
 & & & $\mathrm{SR}\uparrow$ & $\mathrm{TR}\uparrow$ & $\mathrm{CR}\downarrow$ & $\mathrm{SR}\uparrow$ & $\mathrm{TR}\uparrow$ & $\mathrm{CR}\downarrow$ & $\mathrm{SR}\uparrow$ & $\mathrm{TR}\uparrow$ & $\mathrm{CR}\downarrow$ \\
\hline
\rowcolor{gray!18}
\multicolumn{12}{c}{\textbf{\textit{Multi-camera references}}} \\
ABot-N0$^{*}$~\cite{abotn0} & 4B & \ding{55} & 86.9 & 87.6 & 8.5 & 66.7 & 75.4 & 11.6 & 67.3 & 79.5 & 7.1 \\
CoMaTrack$^{*}$~\cite{comatrack} & 3B & \ding{51} & 92.1 & 90.3 & 0.9 & 74.2 & 80.5 & 2.1 & 57.5 & 73.4 & 12.0 \\
NavFoM$^{*}$~\cite{navfom} & 7B & \ding{55} & 88.4 & 80.7 & -- & 62.0 & 67.9 & -- & -- & -- & -- \\
TrackVLA++~\cite{liu2025trackvlaplusplus} & 7B & \ding{55} & 90.9 & 82.7 & 1.5 & 74.0 & 73.7 & 3.5 & 55.9 & 63.8 & 15.1 \\
\hline
\rowcolor{gray!18}
\multicolumn{12}{c}{\textbf{\textit{Single-view methods}}} \\
IBVS$^{\dagger}$~\cite{ibvs} & -- & \ding{55} & 42.9 & 56.2 & 3.8 & 10.6 & 28.4 & 6.1 & 15.2 & 39.5 & 4.9 \\
PoliFormer$^{\dagger}$~\cite{poliformer} & -- & \ding{51} & 4.7 & 15.5 & 40.1 & 2.6 & 13.2 & 44.5 & 3.0 & 15.4 & 41.5 \\
EVT~\cite{evt} & -- & \ding{51} & 24.4 & 39.1 & 42.5 & 3.2 & 11.2 & 47.9 & 17.4 & 21.1 & 45.6 \\
EVT$^{\ddagger}$~\cite{evt} & -- & \ding{51} & 32.5 & 49.9 & 40.5 & 15.7 & 35.7 & 53.3 & 18.3 & 21.0 & 44.9 \\

Uni-NaVid$^{*}$~\cite{zhang2024uninavid} & 7B & \ding{55} & 53.3 & 67.2 & 12.6 & 31.9 & 50.1 & 21.3 & 15.8 & 41.5 & 26.5 \\
NavFoM$^{*}$~\cite{navfom} & 7B & \ding{55} & 85.0 & 80.5 & -- & 61.4 & 68.2 & -- & -- & -- & -- \\
VLingNav$^{*}$~\cite{vlingnav} & 7B & \ding{51} & 88.4 & 81.2 & 2.1 & 67.7 & 73.5 & 5.5 & -- & -- & -- \\
LightNav-0~\cite{wang2026lightnav} & 4B & \ding{51} & 91.7 & 87.7 & 1.9 & 82.6 & 80.1 & 4.6 & -- & -- & -- \\
Qwen-RobotNav~\cite{qwenrobotnav2026} & 4B & -- & 77.4 & 90.0 & 6.4 & -- & -- & -- & -- & -- & -- \\
Qwen-RobotNav~\cite{qwenrobotnav2026} & 8B & -- & 78.6 & 89.7 & 5.7 & -- & -- & -- & -- & -- & -- \\

OmTrackVLA~\cite{omtrackvla2025} & 1B & \ding{55} & 81.4 & 82.8 & 5.1 & 41.5 & 58.8 & 11.3 & 60.0 & 73.9 & 7.6 \\
TrackVLA~\cite{wang2025trackvla} & 7B & \ding{55} & 85.1 & 78.6 & 1.7 & 57.6 & 63.2 & 5.8 & 50.2 & 63.7 & 17.1 \\
TrackVLA++~\cite{liu2025trackvlaplusplus} & 7B & \ding{55} & 86.0 & 81.0 & 2.1 & 66.5 & 68.8 & 4.7 & 51.2 & 63.4 & 15.9 \\
ReferTrack~\cite{ye2026refertrack} & 4B & \ding{55} & 89.4 & 92.5 & 1.6 & 73.3 & 81.8 & 7.6 & 74.1 & 85.7 & 7.7 \\
\hline
\rowcolor{gray!18}
\multicolumn{12}{c}{\textbf{\textit{Our models}}} \\
Baseline (trajectory only) & 0.8B & \ding{55} & 79.2 & 91.4 & 9.5 & 55.9 & 83.6 & 16.5 & 52.4 & 83.6 & 15.5 \\
DobotTrack & 0.8B & \ding{55} & 90.2 & 92.3 & 3.6 & 72.5 & 83.3 & 8.3 & 70.9 & 85.5 & 6.6 \\
\hline
\end{tabular}
\end{table*}

  \subsection{Real-World Dataset}
  \label{sec:real_dataset}

  \noindent{\fontfamily{lmr}\bfseries\selectfont Real-world robot data collection.}
  To reduce the sim-to-real gap, we construct a real-world human-tracking dataset using a quadruped mobile robot
  operating in indoor offices, outdoor open environments, and public pedestrian areas.
  The dataset contains approximately 10.2 hours of synchronized robot-following data, including 596,669 original
  RGB frames.
  After pose alignment and trajectory preprocessing, these recordings produce 565,994 final training samples.
  The dataset comprises 126 person-tracking episodes.
  We use the rectified front-facing RGB stream at a resolution of $960\times540$.
  Robot-following motions are generated through both human teleoperation and an expert model.
  RGB observations are timestamp-aligned with robot states and poses. Future egocentric trajectories are
  recovered from the aligned robot motion.
  Each sequence is additionally associated with a natural-language description of the designated tracking target,
  including appearance attributes such as clothing, hairstyle, and carried objects.
  The dataset contains 33 distinct target appearance profiles captured under diverse tracking conditions.

  Fig.~\ref{fig:real_world_dataset_overview} summarizes the composition of the real-world dataset and presents
  representative tracking samples with target-centric annotations and future trajectory supervision.

  \noindent{\fontfamily{lmr}\bfseries\selectfont Diverse real-world tracking conditions.}
  The real-world dataset covers four representative tracking conditions:
  indoor single-person tracking, indoor tracking with occlusion, outdoor open-area tracking, and complex outdoor
  street tracking.
  The dataset includes approximately 5.68 hours of indoor recordings and 4.53 hours of outdoor recordings.

  Of these recordings, approximately 1 hour contains nearby or visually similar distractors.
  These recordings introduce challenging cases such as partial occlusion, people crossing paths, temporary target ambiguity, and target re-identification.
  Together with the dedicated occlusion and complex-street scenarios, these data expose the model to realistic visual
  and behavioral variations that are difficult to reproduce faithfully in simulation.

  \noindent{\fontfamily{lmr}\bfseries\selectfont Target-centric annotation.}
  We construct dense target-centric annotations for the real-world recordings using a semi-automatic annotation
  pipeline.
  Qwen3.5~\cite{qwen3.5} is first used to localize the described target and generate a target bounding box, which is then provided as a
  prompt to SAM3~\cite{sam3} to obtain the corresponding segmentation mask.
  The generated annotations are then manually reviewed and corrected to address localization and segmentation errors.

  The final annotations provide 596,323 dense target masks and corresponding bounding boxes.
  Bounding boxes are represented in the same normalized $(c_x,c_y,w,h)$ format as in the simulation data.

  We further derive lightweight structured VSC supervision from the target annotations.
  Horizontal position is categorized as left, center, or right, and apparent target scale as small, medium, or large
  using the same definitions as in the simulation dataset.
  Metric target distance is not reliably available in the real-world recordings.
  Real-world VSC supervision is restricted to horizontal position and apparent scale.



\section{Experiments}
\label{sec:Experiments}

\subsection{Experimental Setup}
\label{sec:experimental_setup}

\noindent{\fontfamily{lmr}\bfseries\selectfont Benchmarks and data.}
We evaluate language-guided person tracking on EVT-Bench~\cite{wang2025trackvla}, using its STT, DT, and AT settings to assess single-target tracking, resistance to distractors, and target disambiguation, respectively. Training uses the simulation and real-world data described in Sec.~\ref{sec:dataset}.

\noindent{\fontfamily{lmr}\bfseries\selectfont Evaluation metrics.}
Following the EVT-Bench protocol~\cite{wang2025trackvla}, we report success rate (SR), tracking rate (TR), and collision rate (CR) separately for each setting. 
SR measures the fraction of episodes ending with the agent facing the target at a distance of $1$--$3$\,m. TR measures the fraction of simulation steps satisfying the benchmark's tracking condition. CR measures the fraction of episodes terminated by a collision with the target person. 
Higher SR and TR and lower CR indicate better performance.

\noindent{\fontfamily{lmr}\bfseries\selectfont Implementation details.}
The policy takes an appearance-based instruction and $H=8$ RGB observations sampled at an interval of $\Delta=0.2$\,s and resized to $448\times448$. It predicts $P=8$ future egocentric waypoints at $10$\,Hz, spanning $0.8$\,s. Each frame supplies $N=256$ visual tokens; the target-conditioned compressor uses three attention blocks to retain $N_q=16$ tokens per historical frame. The target-conditioned ablation holds all settings except guidance fixed; the token-budget comparison evaluates 256-token pooling, 16-token pooling, and 16-token Q-identity. Dense current-frame features remain available to the mask head. We train for about $28{,}000$ optimization steps on eight NVIDIA B300 GPUs
with a global batch size of $256$ ($32$ per GPU), using AdamW and
BF16 mixed precision.
The visual encoder is frozen, while the multimodal projector,
language model, Q-identity, and task heads are jointly optimized.
We use peak learning rates of $2\times10^{-5}$ for the language model
and SAM2 mask decoder, and $4\times10^{-5}$ for the remaining
trainable parameters, with a weight decay of $10^{-3}$.
Learning rates follow a one-cycle schedule with $100$ warmup steps
and cosine annealing.

    \begin{table}[!b]
    \centering
    \caption{Auxiliary-task ablation on DT (\%) with the 0.8B model. All variants include trajectory supervision. }
    \label{tab:auxiliary_results}
    \footnotesize
    \setlength{\tabcolsep}{6pt}
    \renewcommand{\arraystretch}{1.08}
    \begin{tabular}{lrrr}
    \hline
    \multirow{2}{*}{Auxiliary task} & \multicolumn{3}{c}{DT} \\
    \cmidrule(lr){2-4}
    & $\mathrm{SR}\uparrow$ & $\mathrm{TR}\uparrow$ & $\mathrm{CR}\downarrow$ \\
    \hline
    Trajectory only & 55.9 & 83.6 & 16.5 \\
    + Box & 59.6 & 84.6 & 14.7 \\
    + VSC & 59.6 & 83.8 & 13.1 \\
    + Mask & 61.3 & 81.5 & 12.0 \\
    + Box \& Mask & 63.7 & 84.5 & 13.2 \\
    + Future prediction & 60.9 & 80.9 & 13.9 \\
    + DINO Future prediction & 58.9 & 80.6 & 15.9 \\
    \hline
    \end{tabular}
    \end{table}

    \begin{table*}[!t]
    \centering
    \caption{Ablation of auxiliary supervision and target conditioning with the 0.8B model and 16-token Q-identity (\%). The first two rows isolate target conditioning under mask supervision. Bold marks the highest SR values.}
    \label{tab:compression_results}
    \footnotesize
    \setlength{\tabcolsep}{3.2pt}
    \renewcommand{\arraystretch}{1.08}
    \begin{tabular}{lcrrr@{\hspace{10pt}}rrr@{\hspace{10pt}}rrr}
    \hline
    \multirow{2}{*}{Auxiliary tasks} & \multirow{2}{*}{Target conditioned} & \multicolumn{3}{c}{STT} & \multicolumn{3}{c}{DT} & \multicolumn{3}{c}{AT} \\
    \cmidrule(lr){3-5}\cmidrule(lr){6-8}\cmidrule(lr){9-11}
    & & $\mathrm{SR}\uparrow$ & $\mathrm{TR}\uparrow$ & $\mathrm{CR}\downarrow$ & $\mathrm{SR}\uparrow$ & $\mathrm{TR}\uparrow$ & $\mathrm{CR}\downarrow$ & $\mathrm{SR}\uparrow$ & $\mathrm{TR}\uparrow$ & $\mathrm{CR}\downarrow$ \\
    \hline
    Mask & \ding{55} & 79.2 & 90.6 & 13.5 & 55.3 & 79.0 & 22.5 & 57.7 & 84.0 & 17.6 \\
    Mask & \ding{51} & 84.8 & 90.0 & 4.5 & 67.6 & 82.0 & 10.3 & 65.6 & 83.6 & 7.6 \\
    Mask \& Box & \ding{51} & \textbf{87.6} & 91.4 & 5.6 & \textbf{69.0} & 83.4 & 13.1 & \textbf{67.8} & 85.6 & 11.5 \\
    \hline
    \end{tabular}
    \end{table*}

    \begin{table*}[!t]
    \centering
    \caption{Representation and token-budget comparison with the 0.8B mask-supervised model. Avg. SR is the mean over the three tasks. Bold marks the highest SR values. We report the 95th-percentile (P95) inference latency measured on Orin Nano.}
    \label{tab:token_results}
    \footnotesize
    \setlength{\tabcolsep}{2.8pt}
    \renewcommand{\arraystretch}{1.08}
    \begin{tabular}{lcrrr@{\hspace{10pt}}rrr@{\hspace{10pt}}rrr@{\hspace{10pt}}cr}
    \hline
    \multirow{2}{*}{Compression} & \multirow{2}{*}{Tokens} & \multicolumn{3}{c}{STT} & \multicolumn{3}{c}{DT} & \multicolumn{3}{c}{AT} & \multirow{2}{*}{Avg. SR $\uparrow$}  & \multirow{2}{*}{\shortstack{P95 latency\\(ms) $\downarrow$}} \\
    \cmidrule(lr){3-5}\cmidrule(lr){6-8}\cmidrule(lr){9-11}
    &  & $\mathrm{SR}\uparrow$ & $\mathrm{TR}\uparrow$ & $\mathrm{CR}\downarrow$ & $\mathrm{SR}\uparrow$ & $\mathrm{TR}\uparrow$ & $\mathrm{CR}\downarrow$ & $\mathrm{SR}\uparrow$ & $\mathrm{TR}\uparrow$ & $\mathrm{CR}\downarrow$ &  & \\
    \hline
    Pooling & 256 & \textbf{86.6} & 89.7 & 4.7 & 61.5 & 77.5 & 13.2 & 59.7 & 82.4 & 12.7 & 69.3  & 203.9 \\
    Pooling & 16 & 84.6 & 88.6 & 6.8 & 61.3 & 81.5 & 12.0 & 57.0 & 81.9 & 11.5 & 67.6  & 91.0 \\
    Q-identity & 16 & 84.8 & 90.0 & 4.5 & \textbf{67.6} & 82.0 & 10.3 & \textbf{65.6} & 83.6 & 7.6 & \textbf{72.7}  & 93.4 \\
    \hline
    \end{tabular}
    \end{table*}

\begin{figure}[t]
    \centering
    \includegraphics[width=\columnwidth]{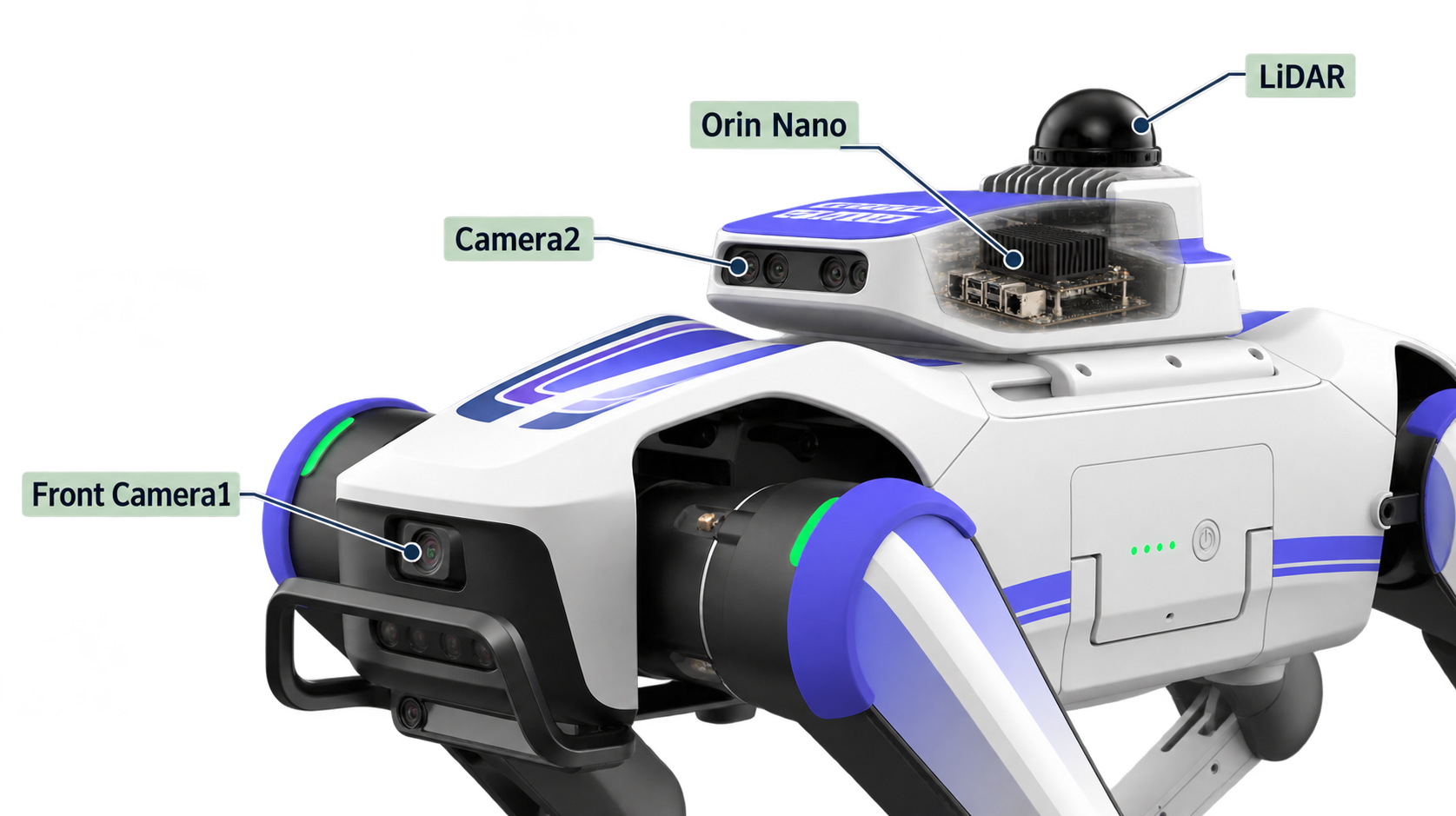}
    \caption{Real-world robotic platform and onboard sensors.}
    \label{fig:robot_hardware}
\end{figure}

\noindent{\fontfamily{lmr}\bfseries\selectfont Real-world deployment.}
The real-world platform is a quadruped robot equipped with front-facing cameras, a LiDAR sensor, and an onboard Orin Nano computer, as shown in Fig.~\ref{fig:robot_hardware}. The tracking policy uses an RGB observation window and a target description to predict egocentric waypoints for the robot's motion controller. 

\subsection{Simulation Benchmark Results}
\label{sec:experimental_simulation_result}

Table~\ref{tab:main_results} compares DobotTrack with published
results on EVT-Bench. Our 0.8B model achieves SRs of
$90.2\%$, $72.5\%$, and $70.9\%$ on STT, DT, and AT,
respectively. Compared with the 1B OmTrackVLA, DobotTrack
raises SR by $8.8$, $31.0$, and $10.9$ percentage points
and records lower collision rates across all three tasks.
These results suggest that target-grounded supervision benefits a compact policy under distractors and appearance ambiguity.

DobotTrack also performs favorably against larger single-view
policies. It exceeds TrackVLA++ by $4.2$, $6.0$, and $19.7$
SR points on STT, DT, and AT, respectively, although its
collision rates remain higher on STT and DT.
Compared with ReferTrack, which uses an external detector,
DobotTrack achieves higher STT SR and lower AT CR, while
its DT and AT SR remain lower. LightNav-0~\cite{wang2026lightnav}, which uses a residual vector-quantized (RVQ) action tokenizer and RL post-training, achieves SRs of $91.7\%$ on STT and $82.6\%$ on DT, exceeding DobotTrack by $1.5$ and $10.1$ percentage points, respectively; AT results are not reported.

Multi-camera methods provide additional references.
Relative to the RL-trained CoMaTrack, DobotTrack achieves
$13.4$ points higher AT SR and $5.4$ points lower AT CR,
while its STT and DT SR remain lower. Because these methods
use different camera inputs and training protocols, these
comparisons provide context rather than controlled evaluations.

\begin{figure*}[!t]
    \centering
    \includegraphics[width=\textwidth]{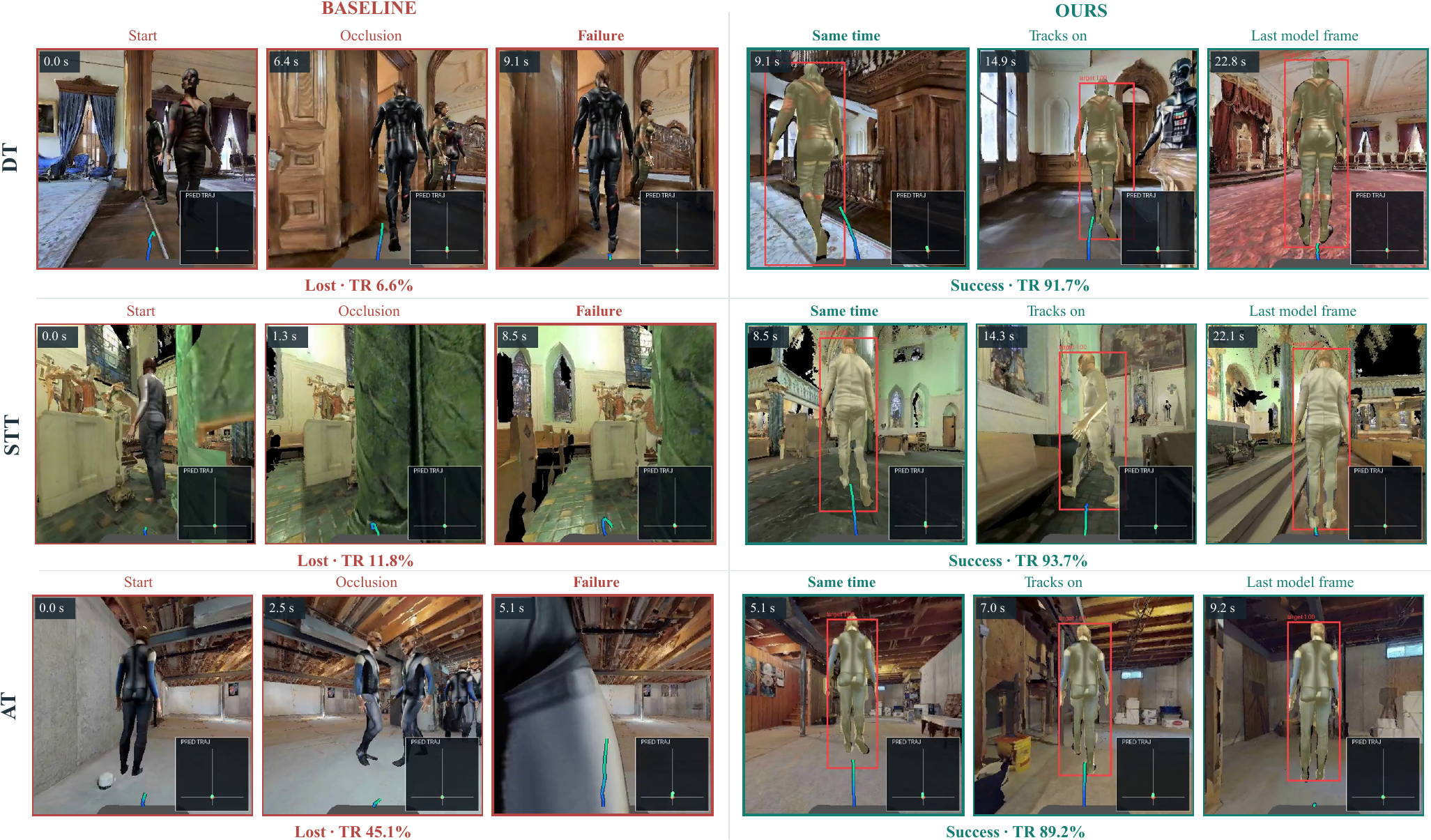}
    \caption{Qualitative tracking comparisons in simulation. Rows show DT, STT, and AT episodes. Left: trajectory-only baseline observations from the start through occlusion to tracking failure. Right: DobotTrack observations at the baseline failure time and subsequent timesteps. Timestamps indicate elapsed episode time; TR values are episode-level results for the illustrated runs.}
    \label{fig:simulation_comparison}
\end{figure*}

Fig.~\ref{fig:simulation_comparison} provides qualitative comparisons on DT, STT, and AT. In these examples, the trajectory-only baseline loses the target under occlusion or interference from other pedestrians, whereas DobotTrack continues tracking the target.

\subsection{Ablation Study}
\label{sec:ablation_study}

We examine the auxiliary supervision, target-conditioned
historical compression, and real-world training data used
in DobotTrack. Unless otherwise stated, experiments use
the 0.8B backbone and 16 tokens
per historical frame.

\noindent{\fontfamily{lmr}\bfseries\selectfont Auxiliary supervision.}
We evaluate five auxiliary tasks providing different forms of spatial
and temporal supervision.
\textit{VSC} classifies spatial attributes of the target person,
including image-plane position (left, center, or right) and distance
to the agent (near, medium, or far).
\textit{Future DINO feature prediction} regresses visual representations
extracted by a frozen DINO encoder from future observations, providing
feature-level supervision for anticipating scene changes.
\textit{Future frame prediction} reconstructs future RGB observations,
providing pixel-level supervision of future visual appearance.
Each task is evaluated as an auxiliary objective alongside trajectory
prediction.
Table~\ref{tab:auxiliary_results} compares auxiliary tasks
for tracking policy learning. Structured VSC improves DT SR
over trajectory-only training, showing that categorical
target attributes provide useful supervision. Box grounding
achieves the same SR as VSC, while mask supervision
yields higher SR and lower CR. Joint mask and box supervision
further raises SR to $63.7\%$, supporting their complementary
roles in learning target representations and spatial localization.

Motivated by latent-space video prediction in DINO-world~\cite{baldassarre2025dinoworld}, we also evaluate DINO future-feature prediction. It improves SR over the trajectory-only baseline, but achieves a lower SR and a higher collision rate than the model trained with joint mask and box supervision. This comparison
suggests that explicit target-level supervision is more
effective than self-supervised future-feature prediction
in our tracking setting.

\begingroup
\makeatletter
\def\ftype@figure{4}
\makeatother
\begin{figure}[!t]
    \centering
    \includegraphics[width=\columnwidth]{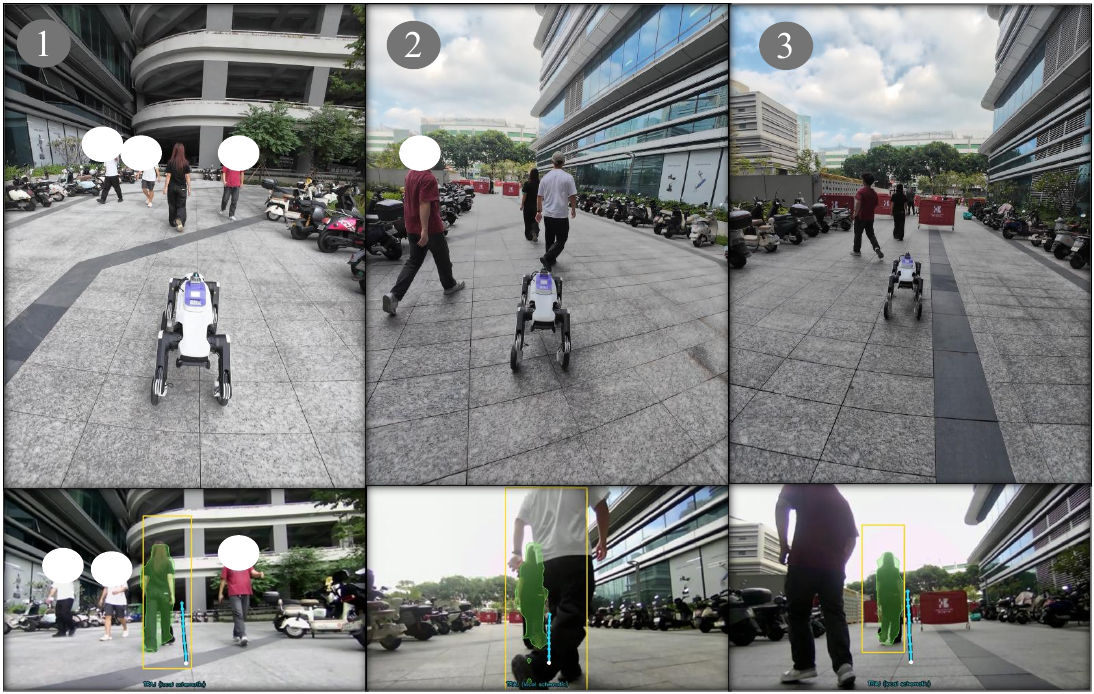}
    \caption{\textbf{Test on real-world system.} DobotTrack follows the designated person in an outdoor pedestrian scene; the predicted mask, bounding box, and egocentric trajectory are overlaid on the robot-centric view.}
    \label{fig:ours_vis}
\end{figure}
\endgroup

\begin{figure*}[!b]
    \centering
    \includegraphics[width=\textwidth]{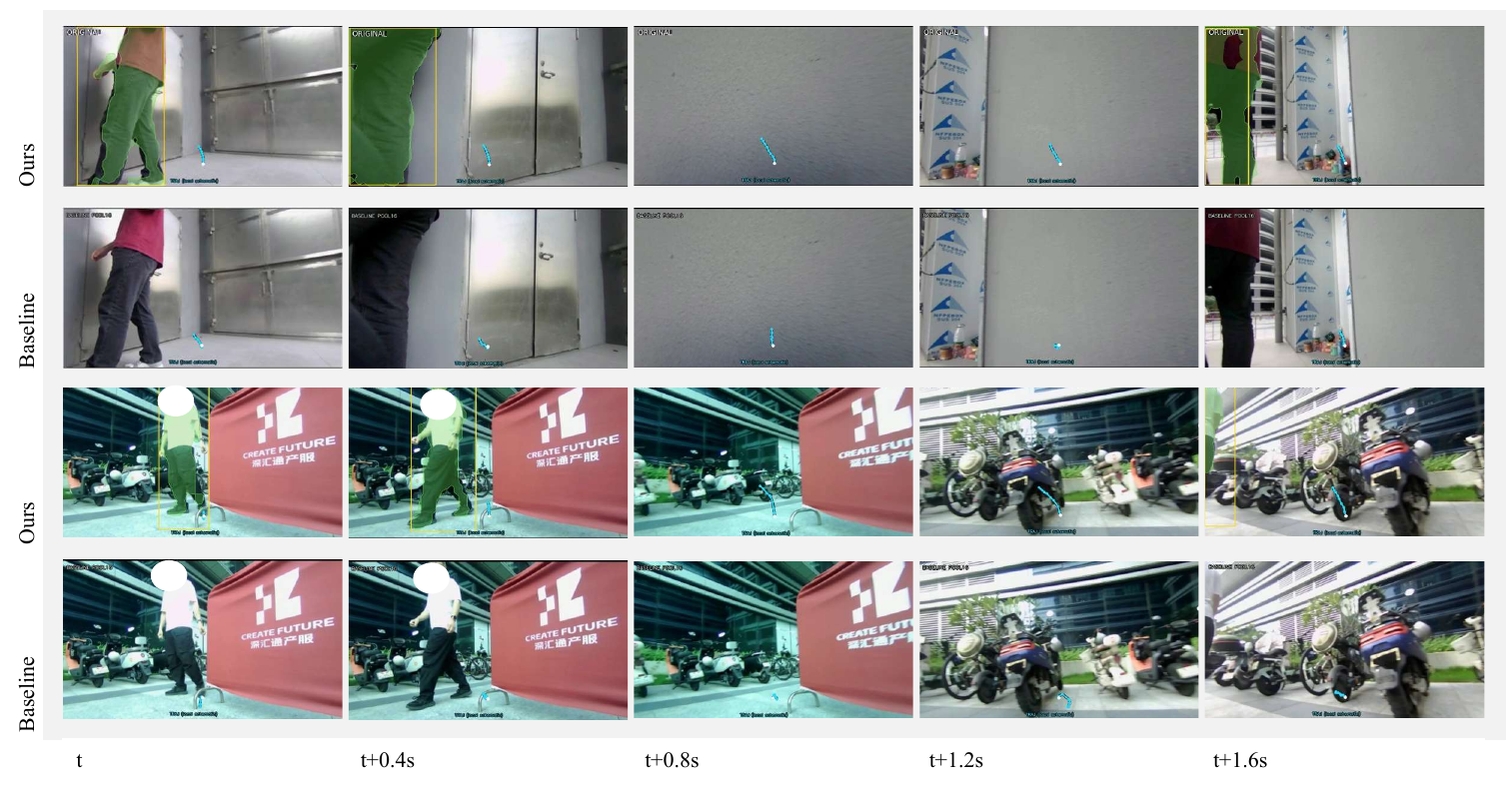}
    \caption{Real-world qualitative comparison between DobotTrack and the trajectory-only baseline in two challenging target-tracking sequences. For each sequence, the upper and lower rows show our method and the baseline, respectively, with frames sampled at 0.4 s intervals. Green masks and yellow bounding boxes indicate target segmentation and localization, while cyan curves represent predicted egocentric trajectories. All visualized predictions are generated by the models. The examples illustrate improved target localization and trajectory prediction under partial visibility and temporary target disappearance.}
    \label{fig:baseline_compare}
\end{figure*}

\noindent{\fontfamily{lmr}\bfseries\selectfont Compressor effectiveness.}
The first two rows of Table~\ref{tab:compression_results}
isolate target conditioning under mask supervision.
Target conditioning raises DT and AT SR by $12.3$ and $7.9$ points,
respectively, and reduces collisions across all three tasks.
These results suggest that target appearance and spatial
priors reduce the burden of identifying relevant features
during historical compression. With joint mask and box
supervision, guided Q-identity reaches SRs of $87.6\%$,
$69.0\%$, and $67.8\%$ on STT, DT, and AT, respectively,
although its collision rates are higher than those of the
mask-supervised guided variant.

Table~\ref{tab:token_results} compares compression methods
and token budgets under mask supervision. At the same
16-token budget, Q-identity maintains similar STT SR
while increasing DT and AT SR by $6.3$ and $8.6$ points,
respectively, and reducing CR across all three tasks.
Compared with 256-token pooling, it achieves a higher
mean SR using one-sixteenth as many historical tokens.

\noindent{\fontfamily{lmr}\bfseries\selectfont Inference efficiency.}
To examine the computational challenge illustrated in Fig.~\ref{fig:tracking_overview} (b), we additionally report deployment latency in Table~\ref{tab:token_results}. With seven historical frames and one current frame, reducing the historical budget from 256 to 16 tokens decreases the visual input from 2048 to 368 tokens. In an optimized deployment on Orin Nano, 256-token pooling, 16-token pooling, and the 16-token target-conditioned compressor have P95 inference latencies of $203.9$, $91.0$, and $93.4$\,ms, respectively. The compact joint architecture is $2.18\times$ faster than 256-token pooling, with a $2.4$\,ms overhead over 16-token pooling.

\noindent{\fontfamily{lmr}\bfseries\selectfont Effectiveness of real-world training data.}
Table~\ref{tab:real_data_results} reports results for the compact model trained with mask and box supervision and Q-identity. The simulation-only model achieves SRs of $87.6\%$, $69.0\%$, and $67.8\%$ on STT, DT, and AT, respectively. With simulation and real-world training data, the corresponding SRs reach $90.2\%$, $72.5\%$, and $70.9\%$. These results suggest that incorporating real-world tracking
data improves tracking success rates in simulated environments.

\begin{table}[!htb]
\centering
\caption{Effect of real-world training data on EVT-Bench SR (\%).}
\label{tab:real_data_results}
\footnotesize
\setlength{\tabcolsep}{6pt}
\renewcommand{\arraystretch}{1.08}
\begin{tabular}{lccc}
\hline
Training data & STT $\uparrow$ & DT $\uparrow$ & AT $\uparrow$ \\
\hline
Simulation only & 87.6 & 69.0 & 67.8 \\
Simulation + real & 90.2 & 72.5 & 70.9 \\
\hline
\end{tabular}
\end{table}


\subsection{Real-World Qualitative Evaluation}

We further evaluate DobotTrack on a quadruped robot
in real-world environments. As shown in Fig.~\ref{fig:ours_vis},
the robot follows the designated person in an outdoor
pedestrian environment while predicting target masks,
bounding boxes, and egocentric trajectories.
These results illustrate the feasibility of deploying
the unified perception-and-action framework on a
physical robot.

Fig.~\ref{fig:baseline_compare} provides qualitative comparisons with the
trajectory-only baseline in two challenging real-world
sequences. Under partial visibility and temporary target
disappearance, DobotTrack maintains more consistent
target localization and trajectory predictions.
These observations are consistent with target-centric supervision and Q-identity:
mask and box supervision strengthen target grounding,
while Q-identity preserves target-specific appearance
and spatial cues from historical observations to support
temporal reasoning when current visual evidence is
limited.

\section{Conclusion}
We presented DobotTrack, a target-grounded vision-language-action framework for language-guided embodied person tracking. The framework combines mask and bounding-box supervision with Q-identity, which uses target-region cues and full-image context to compress historical observations for trajectory prediction. On EVT-Bench, our 0.8B model achieves success rates of $90.2\%$, $72.5\%$, and $70.9\%$ on STT, DT, and AT, respectively. Under mask supervision and the same budget of 16 tokens per historical frame, Q-identity improves DT and AT success rates over pooling by $6.3$ and $8.6$ percentage points, with a $2.4$\,ms increase in P95 inference latency on Orin Nano. We also construct a quadruped-collected dataset comprising 126 person-tracking episodes and 596,323 mask-annotated frames; incorporating these real-world data further improves tracking success on the simulation benchmark. These findings support target-centric supervision and target-conditioned history compression as effective components of compact embodied tracking policies.

The current design relies on the quality of cached grounding predictions and a finite observation window. Future work will investigate robustness to incorrect target associations and prolonged occlusion, and quantify closed-loop performance across a broader range of real-world scenes and target identities.

\FloatBarrier
\section*{Acknowledgments}

We sincerely thank all members of our robotics research team for their dedication, valuable discussions, and contributions throughout this project. In particular, we are deeply grateful to the team members who devoted considerable time and effort to collecting real-world tracking data, often working outdoors for extended periods under the intense summer heat. Their dedication and careful work in data collection and quality control were essential to the construction of our real-world dataset and the experiments presented in this paper.

We also thank Dobot for providing the robotic platform, computational resources, and technical support that made this research possible. We further acknowledge the broader robotics research community, whose advances in vision-language-action models, visual tracking, and embodied navigation have inspired and informed our work.

\bibliographystyle{IEEEtran}

\bibliography{references_revised}

\end{document}